\documentclass[10pt,twocolumn,letterpaper]{article}

\usepackage{cvpr}              %

\usepackage[accsupp]{axessibility}

\usepackage[table,dvipsnames]{xcolor}

\usepackage{tabularx}
\newcolumntype{Y}{>{\centering\arraybackslash}X}
\usepackage{booktabs}
\usepackage{multicol}
\usepackage{multirow}
\usepackage{color}
\usepackage[tableposition=top]{caption}
\usepackage{subcaption}
\usepackage{hhline}
\usepackage{pifont}
\usepackage{threeparttable}
\usepackage{makecell}
\usepackage{algorithm} 
\usepackage{listings}
\usepackage{amsfonts,amssymb}
\usepackage{amsmath}
\usepackage{graphicx}
\usepackage{lipsum}
\usepackage{arydshln}
\usepackage[accsupp]{axessibility}
\usepackage{enumitem}
\usepackage[section]{placeins}

\def\ie{\emph{i.e.}}
\def\eg{\emph{e.g.}}

\def\etal{{\em et al.~}}
\def\vs{{\em v.s.}}

\newcommand{\tabincell}[2]{\begin{tabular}{@{}#1@{}}#2\end{tabular}}

\newcommand{\myPara}[1]{\vspace{.05in}\noindent\textbf{#1.}}
\newlength\savedwidth
\newcommand\whline{\noalign{\global\savedwidth\arrayrulewidth\global\arrayrulewidth 0.8pt}\hline\noalign{\global\arrayrulewidth\savedwidth}}
\usepackage{verbatim}

\definecolor{mylb}{RGB}{229, 247, 255}
\definecolor{darkred}{rgb}{0.55, 0.0, 0.0}
\definecolor{myred}{HTML}{ED058C}
\newcommand{\br}{\rowcolor[RGB]{229, 247, 255}}
\newcolumntype{a}{>{\columncolor{mylb}}c}

\usepackage{blindtext}

\usepackage{amsthm}
\makeatletter
\DeclareRobustCommand{\iscircle}{\mathord{\mathpalette\is@circle\relax}}
\newcommand\is@circle[2]{%
  \begingroup
  \sbox\z@{\raisebox{\depth}{$\m@th#1\bigcirc$}}%
  \sbox\tw@{$#1\square$}%
  \resizebox{!}{\ht\tw@}{\usebox{\z@}}%
  \endgroup
}
\newcommand{\modelname}{InceptionNeXt}

\definecolor{cvprblue}{rgb}{0.21,0.49,0.74}
\usepackage[pagebackref,breaklinks,colorlinks,citecolor=cvprblue]{hyperref}

\def\paperID{6231} %
\def\confName{CVPR}
\def\confYear{2024}

\title{InceptionNeXt: When Inception Meets ConvNeXt}

\author{
Weihao Yu\textsuperscript{1} 
\quad Pan Zhou\textsuperscript{2,3}
\quad Shuicheng Yan\textsuperscript{4}
\quad Xinchao Wang\textsuperscript{1}\thanks{Corresponding Author.}
\\
\textsuperscript{1}{National University of Singapore} ~ \textsuperscript{2}{Singapore Management University} ~ 
\textsuperscript{3}{Sea AI Lab} ~ 
\textsuperscript{4}{Skywork AI}
\\
\small{\texttt{weihaoyu@u.nus.edu} \quad \texttt{panzhou@smu.edu.sg} \quad 
\texttt{shuicheng.yan@kunlun-inc.com} \quad 
\texttt{xinchao@nus.edu.sg}}
\\
\small{Code: \url{https://github.com/sail-sg/inceptionnext}}
}

\begin{document}
\maketitle
\begin{abstract}
Inspired by the long-range modeling ability of ViTs, large-kernel convolutions are widely studied and adopted recently to enlarge the receptive field and improve model performance, like the remarkable work ConvNeXt which employs $7 \times 7$ depthwise convolution. Although such depthwise operator only consumes a few FLOPs, it largely harms the model efficiency on powerful computing devices due to the high memory access costs. For example, ConvNeXt-T has similar FLOPs with ResNet-50 but only achieves $\sim 60\%$ throughputs when trained on A100 GPUs with full precision. Although reducing the kernel size of ConvNeXt can improve speed, it results in significant performance degradation, which poses a challenging problem: How to speed up large-kernel-based CNN models while preserving their performance. To tackle this issue, inspired by Inceptions, we propose to decompose large-kernel depthwise convolution into four parallel branches along channel dimension, \textit{i.e.}, small square kernel, two orthogonal band kernels, and an identity mapping. With this new Inception depthwise convolution, we build a series of networks, namely IncepitonNeXt, which not only enjoy high throughputs but also maintain competitive performance. For instance, InceptionNeXt-T achieves $1.6 \times $ higher training throughputs than ConvNeX-T, as well as attains 0.2\% top-1 accuracy improvement on ImageNet-1K. We anticipate InceptionNeXt can serve as an economical baseline for future architecture design to reduce carbon footprint.

\end{abstract}

\section{Introduction}

\begin{figure}[t]
\vspace{-6mm}
\begin{center}
   \includegraphics[width=1.0\linewidth]{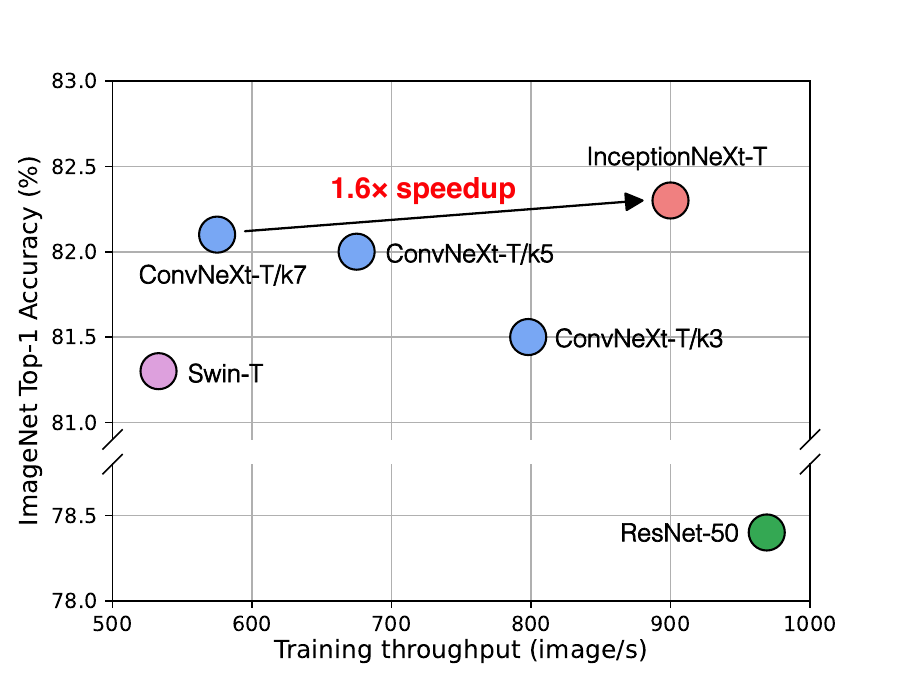}
\end{center}
\vspace{-7mm}
\caption{\textbf{Trade-off between accuracy and training throughput.} All models are trained under the  DeiT training hyperparameters \cite{deit, swin, convnext, resnetsb}. The training throughput is measured on an A100 GPU with batch size of 128. ConvNeXt-T/k$n$ means variants with depthwise convolution kernel size of $n \times n$. \textbf{\modelname{}-T enjoys both ResNet-50's speed and ConvNeXt-T's accuracy.}  }
\label{fig:first_figure}
\vspace{-4mm}
\end{figure}

\begin{figure*}[htp]
\begin{center}
   \includegraphics[width=0.7\linewidth]{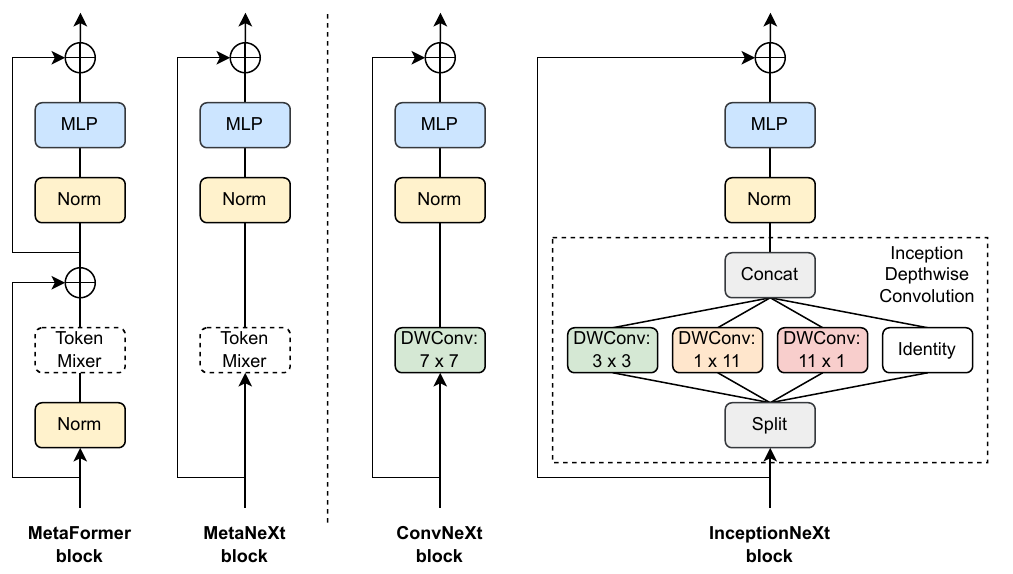}
\end{center}
\vspace{-6mm}
\caption{\textbf{Block illustration of MetaFormer, MetaNext, ConvNeXt and \modelname{}.} Similar to MetaFormer block \cite{metaformer}, MetaNeXt is a general block abstracted from ConvNeXt \cite{convnext}. 
MetaNeXt can be regarded as a simpler version obtained from MetaFormer by merging two residual sub-blocks into one. 
It is worth noting that the token mixer used in MetaNeXt cannot be too complex (\eg,~self-attention \cite{transformer}) or it may fail to train to converge.
By specifying the token mixer as depthwise convolution or Inception depthwise convolution, the model is instantiated as ConvNeXt or InceptionNeXt block. Compared with ConvNeXt, InceptionNeXt is more efficient because it decomposes expensive large-kernel depthwise convolution into four efficient parallel branches.}
\label{fig:block}
\vspace{-5mm}
\end{figure*}

Reviewing the history of deep learning \cite{lecun2015deep}, Convolutional Neural Networks (CNNs) \cite{lecun1989backpropagation, lecun1998gradient} are definitely the most popular models in computer vision. The watershed moment arrived in 2012 when AlexNet \cite{alexnet} claimed victory in the ImageNet contest, ushering in a new era for CNNs in computer vision~\cite{alexnet, imagenet_cvpr, imagenet_ijcv}. Since then, a myriad of influential CNNs has emerged like    Network In Network \cite{lin2013network}, VGG \cite{vgg}, Inception Nets \cite{inception_v1}, ResNe(X)t \cite{resnet, resnext}, DenseNet \cite{huang2017densely} and other efficient models \cite{howard2017mobilenets, sandler2018mobilenetv2, zhang2018shufflenet, tan2019efficientnet, tan2021efficientnetv2}.

Motivated by the great achievement of Transformer in NLP, researchers attempt to integrate its modules or blocks into vision CNN models \cite{wang2018non, detr, huang2019ccnet, bello2019attention}, \eg,~the representative works like Non-local Neural Networks \cite{wang2018non} and DETR \cite{detr}, or even make self-attention as stand-alone primitive \cite{ramachandran2019stand, zhao2020exploring}. Moreover, inspired by the language generative pre-training \cite{gpt1}, Image GPT (iGPT) \cite{imageGPT} treats pixels as tokens and adopts pure Transformer for visual self-supervised learning. However, iGPT faces limitations in handling high-resolution images due to computational costs~\citep{imageGPT}. The breakthrough came with Vision Transformer (ViT)~\citep{vit}, which treats  image patches as tokens, leverages a pure Transformer as the backbone, and has demonstrated remarkable performance in image classification after large-scale supervised image pre-training.

Apparently, the success of ViT \cite{vit} further ignites the enthusiasm for Transformer's application in computer vision. Many ViT variants \cite{deit, t2t, pvt, swin, cswin, focal_transformer, li2022mvitv2}, like DeiT \cite{deit} and Swin \cite{swin}, are proposed and have achieved remarkable performance across a wide range of vision tasks. The superior performance of ViT-like models over traditional CNNs (\eg,~Swin-T's 81.2\% \vs~ResNet-50's 76.1\% on ImageNet \cite{swin, resnet, imagenet_cvpr, imagenet_ijcv}) leads many researchers to believe that Transformers will eventually replace CNNs and dominate the field of computer vision.

It is time for CNN to fight back. With advanced training techniques in DeiT \cite{deit} and Swin \cite{swin}, the work of ``ResNet strikes back"~\cite{resnetsb} shows that the performance of ResNet-50 can rise by 2.3\%, up to 78.4\%. Further, ConvNeXt \cite{convnext} demonstrates that with modern modules like GELU~\cite{gelu} activation and large kernel size similar to attention window size \cite{swin}, CNN models can consistently outperform Swin Transformer \cite{swin} in various settings and tasks. ConvNeXt is not alone: More and more works have shown similar observations \cite{replknet, van, focalnet, rao2022hornet, more_convnets, metaformer_baselines, internimage, hou2022conv2former}, like RepLKNet \cite{replknet} and SLaK \cite{more_convnets}. Among these modern CNN models, the common key feature is the large receptive field that is usually achieved by depthwise convolution \cite{mamalet2012simplifying, chollet2017xception} with large kernel size (\eg,~$7\times7$).

However, despite its small FLOPs, depthwise convolution is actually an ``expensive" operator because it brings high memory access costs and can be a bottleneck on powerful computing devices, like GPUs \cite{shufflenet_v2}. 
Moreover, as observed in \cite{replknet}, larger kernel sizes lead to significantly lower speeds. 
 As shown in Figure \ref{fig:first_figure}, the ConvNeXt-T with a default $7\times7$ kernel size is $1.4 \times$ slower than that with small kernel size of $3\times3$, and is $1.8 \times $  slower than ResNet-50, although they have similar FLOPs. However, using a smaller kernel size limits the receptive field, which can result in performance degradation. For example, ConvNeXt-T/k3 suffers a performance drop of $0.6\%$ top-1 accuracy on the ImageNet-1K dataset when compared to ConvNeXt-T/k7, where k$n$ denotes a kernel size of $n \times n$.

This poses a challenging problem: How to speed up large-kernel CNNs while preserving their performance? In this paper, we aim to address this issue by building upon ConvNeXt as our baseline and improving the depthwise convolution module.  Through our preliminary experiments based on ConvNeXt (see Table \ref{tab:pre_exp}),
we find that not all input channels need to undergo the computationally expensive depthwise convolution operation \cite{shufflenet_v2}. Accordingly, we propose to leave some channels unaltered and process only a portion of the channels with the depthwise convolution operation. 
Next, we propose to decompose large kernel of depthwise convolution into several groups of small kernels in Inception style \cite{inception_v1, inception_v3, inception_v4}. Specifically, for the processing channels, $1/3$ of channels are conducted with kernel of $3\times3$,  another $1/3$ are with $1\times k$, and the remaining $1/3$ are with $k \times 1$. With this new simple and cheap operator, termed as ``\textit{Inception depthwise convolution}", our built model \textit{\modelname{}} achieves a much better trade-off between accuracy and speed. For example, as shown in Figure \ref{fig:first_figure}, \modelname{}-T achieves higher accuracy than ConvNeXt-T while enjoying $1.6 \times$ speedup of training throughput similar to ResNet-50.

The contributions of this paper are two-fold. Firstly, we identify the speed bottleneck of ConvNeXt as shown in Figure \ref{fig:first_figure}. To solve this speed bottleneck while keeping accuracy, we propose Inception depthwise convolution which decomposes the expensive depthwise convolution into three convolution branches with small kernel sizes as well as a branch of identity mapping. Secondly, extensive experiments on image classification and semantic segmentation show a better speed-accuracy trade-off of our model \modelname{}  than  ConvNeXt. We hope that \modelname{} can serve as a new CNN baseline to speed up the research of neural architecture design.

\section{Related work}
\vspace{-1mm}
\subsection{Transformer \vs CNN}
\vspace{-1mm}
Transformer \cite{transformer} was introduced in 2017 for NLP tasks because of its parallel training and also better performance than  LSTM. 
Then many famous NLP models are built on Transformer, including GPT series \cite{gpt1, gpt2, gpt3, ouyang2022training}, BERT \cite{bert}, T5 \cite{raffel2020exploring}, and OPT \cite{zhang2022opt}. For the application of the Transformer in vision tasks, Vision Transformer (ViT) is definitely the seminal work, showing that Transformer can achieve impressive performance after large-scale supervised training. Follow-up works \cite{deit, t2t, pvt, tnt, pvtv2, ren2022shunted, ren2023sg} like Swin \cite{swin} continually improve model performance, achieving new state-of-the-art on various vision tasks. These results seem to tell us ``Attention is all you need" \cite{transformer}. 

But it is not that simple. ViT variants like DeiT usually adopt modern training procedures including various advanced techniques of data augmentation \cite{randaugment, autoaugment, mixup, cutmix, random_erasing}, regularization \cite{inception_v3, stochastic_depth} and optimizers \cite{adam, adamw}. Wightman \etal find that with similar training procedures, the performance of ResNet can be largely improved. Besides, Yu \etal \cite{metaformer} argue that the general architecture instead of attention plays a key role in model performance. Han \etal \cite{han2021connection} find by replacing attention in Swin with regular or dynamic depthwise convolution, the model can also obtain comparable performance. ConvNeXt \cite{convnext}, a remarkable work, modernizes ResNet into an advanced version with some designs from ViTs, and the resulting models consistently outperform Swin \cite{swin}. Other works like RepLKNet \cite{replknet}, VAN \cite{van}, FocalNets \cite{focalnet}, HorNet \cite{rao2022hornet}, SLKNet \cite{more_convnets}, ConvFormer \cite{metaformer_baselines}, Conv2Former \cite{hou2022conv2former}, and InternImage \cite{internimage} constantly improve performance of CNNs. 
Despite the high performance obtained, 
these models neglect efficiency, exhibiting lower speed than ConvNeXt. Actually, ConvNeXt is also not an efficient model compared with ResNet. We argue that CNN models should keep the original advantage of efficiency. Thus, in this paper, we aim to improve the model efficiency of CNNs while maintaining high performance.

\subsection{Convolution with large kernels.}
Well-known works, like AlexNet \cite{alexnet} and Inception v1 \cite{inception_v1} already utilize large kernels up to $11 \times 11$ and $7\times 7$, respectively. To improve the efficiency of large kernels, VGG \cite{vgg} proposes to heavily stack $3\times 3$ convolutions while Inception v3 \cite{inception_v3} factorizes $k \times k$ convolution into $1 \times k$ and $k \times 1$ staking sequentially. For depthwise convolution, MixConv \cite{tan2019mixconv} splits kernels into several groups from $3\times 3$ to $k \times k$. Besides, Peng \etal find that large kernels are important for semantic segmentation and they decompose large kernels similar to Inception v3 \cite{inception_v3}. Witnessing the success of Transformer in vision tasks \cite{vit, pvt, swin}, large-kernel convolution is more emphasized since it can offer a large receptive field to imitate attention \cite{han2021connection, convnext}. For example, ConvNeXt adopts kernel size of $7 \times 7 $ for depthwise convolution by default.
To employ larger kernels, RepLKNet \cite{replknet} proposes to utilize structural re-parameterization techniques \cite{zagoruyko2017diracnets, repvgg} to scale up kernel size to $31 \times 31$; VAN \cite{van} sequentially stacks large-kernel depth-wise convolution (DW-Conv) and depth-wise dilation convolution to obtain $21 \times 21$ receptive filed; FocalNets \cite{focalnet} employ a gating mechanism to fuse multi-level features from stacking depthwise convolutions; 
SegNeXt \cite{guo2022segnext} learns multi-scale features by multiple branches of staking $1\times k$ and $k \times 1$. 
Recently, SLaK \cite{more_convnets} factorizes large kernel $k \times k$ into two small non-square kernels ($k \times s$ and $s \times k$ with $s < k$). Unlike these works, we do not aim to scale up larger kernels. Instead, we target efficiency and decompose large kernels in a simple and speed-friendly way while keeping comparable performance.

\vspace{-2mm}
\section{Formulation and Method}
\begin{algorithm}[t]
\vspace{-1mm}
\caption{Inception Depthwise Convolution (PyTorch-like Code)}
\label{alg:code}
\definecolor{codeblue}{rgb}{0.25,0.5,0.5}
\definecolor{codekw}{rgb}{0.85, 0.18, 0.50}
\lstset{
  backgroundcolor=\color{white},
  basicstyle=\fontsize{7.5pt}{7.5pt}\ttfamily\selectfont,
  columns=fullflexible,
  breaklines=true,
  captionpos=b,
  commentstyle=\fontsize{7.5pt}{7.5pt}\color{codeblue},
  keywordstyle=\fontsize{7.5pt}{7.5pt}\color{codekw},
}
\begin{lstlisting}[language=python]
import torch.nn as nn

class InceptionDWConv2d(nn.Module):
    def __init__(self, in_channels, square_kernel_size=3, band_kernel_size=11, branch_ratio=1/8):
        super().__init__()
        
        gc = int(in_channels * branch_ratio) # channel number of a convolution branch
        
        self.dwconv_hw = nn.Conv2d(gc, gc, square_kernel_size, padding=square_kernel_size//2, groups=gc)
        
        self.dwconv_w = nn.Conv2d(gc, gc, kernel_size=(1, band_kernel_size), padding=(0, band_kernel_size//2), groups=gc)
        
        self.dwconv_h = nn.Conv2d(gc, gc, kernel_size=(band_kernel_size, 1), padding=(band_kernel_size//2, 0), groups=gc)
        
        self.split_indexes = (gc, gc, gc, in_channels - 3 * gc)
        
    def forward(self, x):
        # B, C, H, W = x.shape
        x_hw, x_w, x_h, x_id = torch.split(x, self.split_indexes, dim=1)
        
        return torch.cat(
            (self.dwconv_hw(x_hw), 
            self.dwconv_w(x_w), 
            self.dwconv_h(x_h), 
            x_id), 
            dim=1)
\end{lstlisting}
\vspace{-2mm}
\end{algorithm}

\vspace{-3mm}
\subsection{MetaNeXt}
\vspace{-1mm}
\myPara{Formulation of MetaNeXt Block}
In ConvNeXt \cite{convnext}, for its each ConvNeXt block, the input $X$ is first processed by a depthwise convolutioin to propagate information along spatial dimensions. 
We follow MetaFormer \cite{metaformer} to abstract the depthwise convolution as a \textit{token mixer} which is responsible for spatial information interaction. Accordingly, as shown in the second subfigure in Figure \ref{fig:block}, the ConvNeXt  is abstracted as \textit{MetaNeXt} block. Formally, in a MetaNeXt block, its  input $X$ is firstly processed as
\begin{equation}
    X' = \mathrm{TokenMixer}(X),
\end{equation}
where $X, X' \in \mathbb{R}^{B \times C \times H \times W}$ with $B$, $C$, $H$ and $W$ respectively denoting batch size, channel number, height and width. 
Then the output from the token mixer is normalized
\begin{equation}
    Y = \mathrm{Norm}(X').
\end{equation}
After normalization \cite{batch_norm, layer_norm}, the features are then fed into an MLP module which consists of two fully-connected layers with an activation function between them, the same as feed-forward network in Transformer \cite{transformer}. The two fully-connected layers can also be implemented by $1 \times 1$ convolutions. Also, shortcut connection \cite{resnet, highway} is adopted. This process can be expressed by
\begin{equation}
     Y = \mathrm{Conv}_{1 \times 1}^{rC\rightarrow C}\{\sigma[\mathrm{Conv}_{1 \times 1}^{C \rightarrow rC}(Y)]\} + X,
\end{equation}
where $\mathrm{Conv}_{k \times k}^{C_i \rightarrow C_o}$ means convolution with kernel size of $k \times k$, input channels of $C_i$ and output channels of $C_o$; $r$ is the expansion ratio and $\sigma$ denotes activation function.

\myPara{Comparison to MetaFormer block} As shown in Figure \ref{fig:block}, it can be found that MetaNeXt block shares similar modules with MetaFormer block \cite{metaformer}, \eg,~token mixer and MLP. Nevertheless, a critical difference between the two models lies in the number of shortcut connections \cite{resnet, highway}. MetaNeXt block implements a single shortcut connection, whereas the MetaFormer block incorporates two, one for the token mixer and the other for the MLP.
From this aspect, MetaNeXt block can be regarded as a result of merging two residual sub-blocks from MetaFormer, thereby simplifying the overall architecture.
As a result, the MetaNeXt architecture exhibits a higher speed compared to MetaFormer. 
However, this simpler design comes with a limitation: the token mixer component in MetaNeXt cannot be complicated (\eg, Attention) as shown in our experiments (Table \ref{tab:iso}).

\myPara{Instantiation to ConvNeXt} As shown in Figure \ref{fig:block}, in ConvNeXt, the token mixer is simply implemented by a depthwise convolution
\begin{equation}
    X' = \mathrm{TokenMixer}(X) = \mathrm{DWConv}_{k \times k}^{C\rightarrow C}(X),
\end{equation}
where $\mathrm{DWConv}_{k \times k}^{C \rightarrow C}$ denotes depthwise convolution with kernel size of $k \times k$. In ConvNeXt, $k$ is set as 7 by default.

\subsection{Inception depthwise convolution}
\begin{table}[h]
\vspace{-2mm}
\setlength{\tabcolsep}{3pt}
\footnotesize
\centering
\begin{tabular}{c c c c c c c }
\whline
	\multirow{2}{*}{\makecell[c]{Kernel size \\ of DWConv}} & \multirow{2}{*}{\makecell[c]{Convolution \\ ratio}} & \multirow{2}{*}{\makecell[c]{Params \\ (M)}} & \multirow{2}{*}{\makecell[c]{MACs \\ (G)}} & \multicolumn{2}{c}{Throughput} & \multirow{2}{*}{\makecell[c]{Top-1 \\ (\%)}} \\
 ~ & ~ & ~ & ~ & Train & Inference & ~ \\
\whline
 $7\times 7$ & 1.0 & 28.6 & 4.5 & 575 & 2413 & 82.1* \\
 $5 \times 5$ & 1.0 & 28.4 & 4.4 & 675 & 2704 & 82.0 \\
 $3 \times 3$ & 1.0 & 28.3 & 4.4 & 798 & 2802 & 81.5 \\
 $3 \times 3$ & $1/2$ & 28.3 & 4.4 & 818 & 2740 & 81.4 \\
$3 \times 3$ & $3/8$ & 28.3 & 4.4 & 847 & 2762 & 81.4 \\
$3 \times 3$ & $1/4$ & 28.3 & 4.4 & 871 & 2808 & 81.3 \\
$3 \times 3$ & $1/8$ & 28.3 & 4.4 & 901 & 2833 & 80.8 \\
$3 \times 3$ & $1/16$ & 28.3 & 4.4 & 916 & 2846 & 80.1 \\
 
\whline
\end{tabular}

\caption{\textbf{Preliminary experiments based on ConvNeXt-T. } Convolution ratio means the ratio of channels to be processed by depthwise convolution while the other channels keep unchanged. Throughputs are measured on an A100 GPU with batch size of 128 and TF32. * The result is reported in ConvNeXt paper \cite{convnext}.
\label{tab:pre_exp}
}
\vspace{-5mm}
\end{table}
\myPara{Preliminary experiments on ConvNeXt-T}
We first conducted preliminary experiments based on ConvNeXt-T and report the results in Table \ref{tab:pre_exp}. Firstly, the kernel size of depthwise convolution is reduced from $7 \times 7$ to $3 \times 3$. Compared to the model with kernel size of $7 \times 7$, the one with kernel size of $3 \times 3$ enjoys $1.4 \times$ higher training throughput, but suffers a significant performance drop from 82.1\% to 81.5\%. Next, inspired by ShuffleNet V2 \cite{shufflenet_v2}, we only feed partial input channels into depthwise convolution while the remaining ones keep unchanged. The number of processed input channels is controlled by a ratio. It is found that when the ratio is reduced from 1 to $1/4$, the training throughput can be further improved while the performance almost maintains. In summary, these preliminary experiments convey two findings on ConvNeXt. \underline{Finding 1}: Large-kernel depthwise convolution is the speed bottleneck. \underline{Finding 2}: Processing partial channels is good enough in single depthwise convolution layer \cite{shufflenet_v2}. 

\begin{table}
\vspace{-3mm}
\begin{center}
\footnotesize
\setlength{\tabcolsep}{10pt}
\begin{tabular}{l|c|c}
\whline
Conv. type &  Params & FLOPs \\
\whline
Conventional conv. & $k^2C^2$ & $2k^2C^2HW$\\
Depthwise conv. & $k^2C$ & $2k^2CHW$\\
Inception dep. conv. & $(2k+9)C/8$ & $(2k+9)CHW/4$ \\
\whline
\end{tabular}
\end{center}
\vspace{-2mm}
\caption{\textbf{Complexity of different types of convolution.} For simplicity, assume input and output channels are the same, and the bias term is omitted. $k$, $C$, $H$ and $W$ denote kernel size, channel number, height and width, respectively. The parameters and FLOPs of vanilla convolution and depthwise convolution are quadratic to kernel size $k$. In contrast, Inception depthwise convolution is linear to $k$.}
\label{tab:complexity}
\vspace{-3mm}
\end{table}

\begin{figure}[t]
\vspace{-2mm}
\begin{center}
   \includegraphics[width=0.8\linewidth]{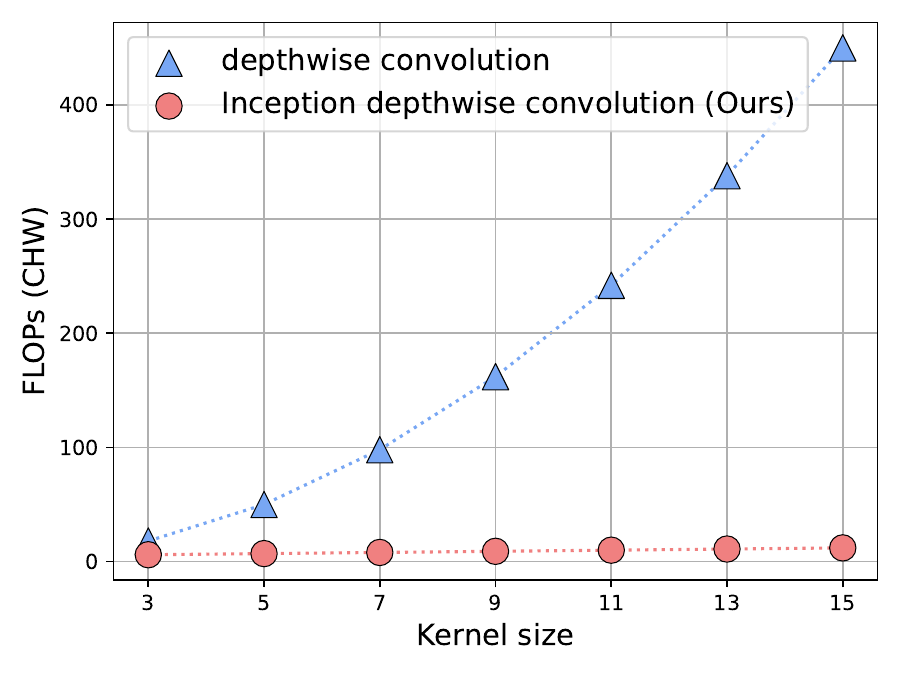}
\end{center}
\vspace{-8mm}
\caption{\textbf{Comparison of FLOPs between depthwise convolution and Inception depthwise convolution.} Inception depthwise convolution is much more efficient than depthwise convolution as kernel size increases.}
\label{fig:two_types}
\vspace{-2mm}
\end{figure}

\begin{table}[t]
\footnotesize
\centering
\setlength{\tabcolsep}{1pt}
\newcommand{\poollayer}{
Band kernel size & 9 & \multicolumn{3}{c}{11}\\
\cline{4-8}
 & & & Conv. group ratio & 1/4 & \multicolumn{3}{c}{1/8} \\
\cline{4-8}
}

\newcommand{\stitle}[7]{
\multirow{5}{*}{#1} & \multirow{5}{*}{\scalebox{1}{$\frac{H}{#2}\times \frac{W}{#2}$}} & \multirow{2}{*}{\tabincell{c}{Down- \\ sampling}} & Kernel Size & \multicolumn{4}{c}{$#3 \times #3$, stride $#4$} \\
\cline{4-8}
    &    &    & Embed. Dim. & \multicolumn{1}{c|}{$#5$} & \multicolumn{2}{c|}{$#6$} & $#7$  \\
\cline{3-8}
& & \multirow{4}{*}{\tabincell{c}{\modelname{}\\Block}} 
}

\begingroup
\renewcommand{\arraystretch}{1.1}
\begin{tabular}{c|c|c|c|c|c|c|c}
\whline
  \multirow{2}{*}{Stage} & \multirow{2}{*}{\#Tokens} & \multicolumn{2}{c|}{\multirow{2}{*}{Layer Specification}} & \multicolumn{4}{c}{\modelname{}} \\
\cline{5-8}
 & & \multicolumn{2}{c|}{} & A & T & S & B \\
\whline
\stitle{1}{4}{4}{4}{40}{96}{128}    & \poollayer
 & & & MLP Ratio & \multicolumn{4}{c}{4} \\
\cline{4-8}
 & & & \# Block & 2 & \multicolumn{3}{c}{3} \\
\hline
\stitle{2}{8}{2}{2}{90}{192}{256}  & \poollayer
 & & & MLP Ratio & \multicolumn{4}{c}{4} \\
 \cline{4-8}
 & & & \# Block & 2 & \multicolumn{3}{c}{3} \\
\hline
\stitle{3}{16}{2}{2}{180}{384}{512} & \poollayer
 & & & MLP Ratio & \multicolumn{4}{c}{4} \\
\cline{4-8}
 & & & \# Block & 6 & 9 & \multicolumn{2}{c}{27} \\
\hline
\stitle{4}{32}{2}{2}{320}{768}{1024} & \poollayer
 & & & MLP Ratio & \multicolumn{4}{c}{3} \\
 \cline{4-8}
 & & & \# Block & 2 & \multicolumn{3}{c}{3} \\
\hline
\multicolumn{8}{c}{Global average pooling, MLP} \\
\hline
\multicolumn{4}{c|}{Parameters~(M)} & 4.2 & 28.1  &  49.4 &  86.7 \\
\hline
\multicolumn{4}{c|}{MACs~(G)} & 0.5 & 4.2 & 8.4 &  14.9 \\
\whline
\end{tabular}
\endgroup

\caption{\textbf{ Configurations of \modelname{} models} which have similar model configurations to ConvNeXt \cite{convnext}. ``A'', ``T'', ``S'' and ``B'' represent ``Atto'', ``Tiny'', ``Small'' and ``Base'', respectively.
}
\label{tab:model}
\vspace{-9mm}
\end{table}

\myPara{Formulation}
Based on the above findings, we propose a new type of convolution to keep both accuracy and efficiency. According to \underline{Fingding 2}, we leave partial channels unchanged and denote them as a branch of identity mapping. Motivated by \underline{Fingding 1}, for the processing channels, we propose to decompose the depthwise operations in Inception style \cite{inception_v1, inception_v3, inception_v4}. 
Inception \cite{inception_v1} utilizes several branches of small kernels (\eg,~$3 \times 3$) and large kernels (\eg,~$5 \times 5$). Similarly, we adopt $3 \times 3$ as one of our branches but get rid of the usage of the large square kernels because of their slow practical speed.  Instead, large kernel $k_h \times k_w$ is decomposed as $1 \times k_w$ and $k_h \times 1$ inspired by Inception v3 \cite{inception_v3}. 

Specifically, for input $X$, we split it into four groups along the channel dimension,
\begin{equation}
\begin{split}
X_\mathrm{hw}, X_\mathrm{w}, X_\mathrm{h}, X_\mathrm{id} &= \mathrm{Split}(X) \\
&= X_{:, :g}, X_{:, g:2g}, X_{:, 2g:3g}, X_{:, 3g:} ,
\end{split}
\end{equation}
where $g$ is the channel numbers of convolution branches. We can set a ratio $r_g$ to determine the branch channel numbers by $g = r_g C$. Next, the splitting inputs are fed into different parallel branches,
\begin{equation}
\begin{split}
X'_\mathrm{hw} &= \mathrm{DWConv}_{k_s \times k_s}^{g\rightarrow g}(X_\mathrm{hw}), \\
X'_\mathrm{w} &= \mathrm{DWConv}_{1\times k_b}^{g\rightarrow g}(X_\mathrm{w}), \\
X'_\mathrm{h} &= \mathrm{DWConv}_{k_b\times 1}^{g\rightarrow g}(X_\mathrm{h}), \\
X'_\mathrm{id} &= X_\mathrm{id}, \\
\end{split}
\end{equation}
where $k_s$ denotes  the small square kernel size set as 3 by default;   $k_b$ represents the band kernel size set as 11 by default. Finally, the outputs from each branch are concatenated,
\begin{equation} \label{eq1}
X' = \mathrm{Concat}(X'_\mathrm{hw}, X'_\mathrm{w}, X'_\mathrm{h}, X'_\mathrm{id}).
\end{equation}
The illustration of \modelname{} block is shown in Figure \ref{fig:block}. Moreover,  its  PyTorch \cite{pytorch} code is summarized  in Algorithm \ref{alg:code}.

\myPara{Complexity} The complexity of three types of convolution, \ie, conventional, depthwise, and Inception depthwise convolution is shown in Table \ref{tab:complexity}. As can be seen, Inception depthwise convolution is much more efficient than the other two types of convolution in terms of parameter numbers of FLOPs. Inception depthwise convolution consumes parameters and FLOPs linear to both channel and kernel size. The comparison of depthwise and Inception depthwise convolutions regarding FLOPs is also clearly shown in Figure \ref{fig:two_types}.

\vspace{-1mm}
\subsection{\modelname{}}
\vspace{-1mm}
Based on InceptionNeXt block, we can build a series of models named InceptionNeXt. Since ConvNeXt \cite{convnext} is the our main comparing baseline, we mainly follow it to build models with several sizes \cite{rw2019timm}. Specifically, similar to ResNet \cite{resnet} and ConvNeXt, \modelname{} also adopts 4-stage framework.  The same as ConvNeXt, the numbers of 4 stages are [2, 2, 6, 2] for atto size,  [3, 3, 9, 3] for small size and [3, 3, 27, 3] for base size. 
We adopt Batch Normalization since this paper emphasizes speed.  Another difference with ConvNeXt is that \modelname{} uses an MLP ratio of 3 in stage 4 and moves the saved parameters to the classifier, which can help reduce a few FLOPs (\eg,~3\% for base size). The detailed model configurations are reported in Table \ref{tab:model}.

\section{Experiment}
\subsection{Image classification}
\vspace{-2mm}

\begin{table*}[t]
    \centering
    \setlength{\tabcolsep}{5pt}
    \scalebox{0.8}{\begin{tabular}{l | c | c | c c | llll | l}
\whline
\multirow{3}{*}{\makecell[c]{Model}}   &  \multirow{3}{*}{\makecell[c]{Mixing \\ Type}}     & 
\multirow{3}{*}{\makecell[c]{Image \\ size}} & \multirow{3}{*}{\makecell[c]{Params \\ (M)}}  & \multirow{3}{*}{\makecell[c]{MACs \\ (G)}} & \multicolumn{4}{c|}{Throughput (img/s)} & \multirow{3}{*}{\makecell[c]{Top-1 \\ (\%)}}
\\
~ & ~ & ~ & ~ & ~ & \multicolumn{2}{c}{A100} & \multicolumn{2}{c|}{2080Ti} \\
~ &  ~ & ~ & ~ & ~ & Train & Infer & Train & Infer & ~ \\
\whline
MobileNetV2 (1.4) \cite{sandler2018mobilenetv2} & Conv & $224^2$ & 6.1 & 0.60 & 1001 & 5190 & 471 & 1859 & 74.7 \\
EfficientNet-B0 \cite{tan2019efficientnet} & Conv & $224^2$ & 5.3 & 0.40 & 954 & 5502 & 464 & 1944 & 77.1 \\
GhostNet 1.3$\times$ \cite{han2020ghostnet} & Conv & $224^2$ & 7.3 & 0.24 & 946 & 7451 & 589 & 2757 & 75.7 \\
\br
ConvNeXt-A \cite{convnext, rw2019timm} & Conv & $224^2$ & 3.7 & 0.55 &  835 & 4539 & 345 & 1568 & 75.7 \\
\br
InceptionNeXt-A (Ours) & Conv & $224^2$ & 4.2 & 0.51 & 2661$_{+219\%}$ & 9876$_{+118\%}$ & 992$_{+188\%}$ & 3595$_{+129\%}$ & 75.3$_{-0.4}$ \\
\hline
DeiT-S \cite{deit}  & Attn  & $224^2$ & 22 & 4.6 & 1227 & 3781 & 276 & 784 & 79.8 \\
T2T-ViT-14 \cite{t2t}  & Attn   & $224^2$ & 22 & 4.8 & -- & -- & -- & -- & 81.5 \\
TNT-S \cite{tnt} & Attn & $224^2$ & 24 & 5.2 & -- & -- & -- & -- & 81.5 \\
Swin-T \cite{swin}  & Attn   & $224^2$ &  29 & 4.5 & 564 & 1768 & 184 & 554 & 81.3  \\
Focal-T \cite{focal_transformer} & Attn & $224^2$ & 29 & 4.9 & -- & -- & -- & -- & 82.2 \\
\hdashline
ResNet-50 \cite{resnet, resnetsb}   & Conv   & $224^2$ & 26 & 4.1 & 969 &  3149 & 278 & 977 & 78.4 \\
RSB-ResNet-50 \cite{resnet, resnetsb}   & Conv   & $224^2$ &  26 & 4.1 & 969 &  3149 & 278 & 977 & 79.8 \\
RegNetY-4G \cite{regnet, resnetsb}   & Conv   & $224^2$ & 21 & 4.0 & 670 & 2694 & 222 & 859 &  81.3 \\
FocalNet-T \cite{focalnet} & Conv & $224^2$ & 29 & 4.5 & -- & -- & -- & -- & 82.3 \\
\br 
ConvNeXt-T \cite{convnext}   & Conv   & $224^2$ & 29 &  4.5 & 575 & 2413 & 177 & 590 & 82.1 \\
\br 
\modelname{}-T (Ours) & Conv & $224^2$ & 28 &  4.2 & 901$_{+57\%}$ & 2900$_{+20\%}$ & 254$_{+44\%}$ & 822$_{+39\%}$ & 82.3$_{+0.2}$ \\
\whline
T2T-ViT-19 \cite{t2t}  & Attn   & $224^2$ & 39 & 8.5 & -- & -- & -- & -- & 81.9 \\
PVT-Medium \cite{pvt} & Attn & $224^2$ & 44 & 6.7 & -- & -- & -- & -- & 81.2 \\
Swin-S \cite{swin}  & Attn   & $224^2$ & 50 &  8.7 & 359  & 1131 & 109 & 328 &  83.0  \\
Focal-S \cite{focal_transformer} & Attn & $224^2$ & 51 & 9.1 & -- & -- & -- & -- & 83.5 \\
\hdashline
RSB-ResNet-101 \cite{resnet, resnetsb}   & Conv   & $224^2$ &  45 & 7.9 & 620 & 2057 & 168 & 592 & 81.3 \\
RegNetY-8G \cite{regnet, resnetsb}   & Conv   & $224^2$ & 39 &  8.0 & 689 & 1326 & 124 & 480 & 82.1 \\
FocalNet-S \cite{focalnet} & Conv & $224^2$ & 50 & 8.7 & -- & -- & -- & -- & 83.5 \\
\br 
ConvNeXt-S \cite{convnext}   & Conv   & $224^2$ & 50 & 8.7 & 361 & 1535 & 105 & 353 & 83.1 \\
\br 
\modelname{}-S (Ours) & Conv & $224^2$ & 49 & 8.4 & 521$_{+44\%}$ & 1750$_{+14\%}$ & 130$_{+24\%}$ & 447$_{+27\%}$ & 83.5$_{+0.4}$ \\
\whline
DeiT-B \cite{deit}  & Attn  & $224^2$ & 86 & 17.5 & 541 & 1608 & 86 & 259 & 81.8 \\
T2T-ViT-24 \cite{t2t}  & Attn & $224^2$ & 64 &  13.8 & -- & -- & -- & -- & 82.3 \\
TNT-B \cite{tnt} & Attn &  $224^2$ & 66 & 14.1 & -- & -- & -- & -- & 82.9 \\
PVT-Large \cite{pvt} & Attn &  $224^2$ & 62 & 9.8 & -- & -- & -- & -- & 81.7 \\
Swin-B \cite{swin}  & Attn  & $224^2$ & 88 &  15.4 & 271 & 843 & 72 & 223 & 83.5 \\
Focal-B \cite{focal_transformer} & Attn & $224^2$ & 90 & 16.0 & -- & -- & -- & -- & 83.8 \\
\hdashline
RSB-ResNet-152 \cite{resnet, resnetsb}   & Conv  & $224^2$ & 60 & 11.6 & 437 & 1457 & 115 & 415 & 81.8 \\
RegNetY-16G \cite{regnet, resnetsb}   & Conv  & $224^2$ & 84 &15.9 & 322 & 1100 & 76 & 295 & 82.2 \\
RepLKNet-31B \cite{replknet}  & Conv & $224^2$ & 79  & 15.3 & -- & -- & -- & -- &  83.5 \\
FocalNet-B \cite{focalnet} & Conv & $224^2$ & 89 & 15.4 & -- & -- & -- & -- & 83.9 \\
\br 
ConvNeXt-B \cite{convnext}   & Conv   & $224^2$ & 89  & 15.4 & 267 & 1122 & 68 & 236 & 83.8  \\
\br
\modelname{}-B (Ours) & Conv & $224^2$ & 87 & 14.9 & 375$_{+40\%}$ & 1244$_{+11\%}$ & 80$_{+18\%}$ & 287$_{+22\%}$ & 84.0$_{+0.2}$ \\
\hline
DeiT-B \cite{deit} & Attn & $384^2$ & 86 & 55.4 & 131 & 361 & 25 & 73 & 83.1 \\
Swin-B \cite{swin} & Attn & $384^2$ & 88 & 47.1 & 104 & 296 & 21 & 65 & 84.5 \\
\hdashline
RepLKNet-31B \cite{replknet} & Conv & $384^2$ & 79 & 45.1 & -- & -- & -- & -- & 84.8 \\
\br
ConvNeXt-B \cite{convnext} & Conv & $384^2$ & 89 & 45.0 & 95 & 393 & 23 & 79 & 85.1 \\
\br
\modelname{}-B (Ours) & Conv & $384^2$ & 87 & 43.6 & 139$_{+46\%}$ & 428$_{+9\%}$ & 27$_{+17\%}$ & 97$_{+23\%}$ & 85.2$_{+0.1}$ \\
\whline
\end{tabular}

}
    \caption{\label{tab:imagenet}
    \textbf{Performance of models trained on ImageNet-1K.} The throughputs are measured on an A100 GPU (PyTorch 1.13.0 and CUDA 11.7.1) with TF32 (TensorFloat-32), and on a 2080Ti (PyTorch 1.8.1 and CUDA 10.2) with FP32. The batch size for throughput benchmarking is initially set as 128 and is reduced until the GPU can host. The better results of ``Channel First" and ``Channel Last"  memory layouts are reported. }
    \vspace{-9mm}
\end{table*}

\myPara{Setup}
For the image classification task, ImageNet-1K \cite{imagenet_cvpr, imagenet_ijcv} is one of the most commonly-used benchmarks, which contains around 1.3 million images in the training set and 50 thousand images in the validation set. To fairly compare with the widely-used baselines, \eg, Swin \cite{swin} and ConvNeXt \cite{convnext}, we mainly follow the training hyper-parameters from DeiT \cite{deit} without distillation. Specifically, the models are trained by AdamW \cite{adamw} optimizer with a learning rate $lr = 0.001 \times \mathrm{batch size} / 1024$ ($lr=4e-3$ and $\mathrm{batch size} = 4096$ are used in this paper the same as ConvNeXt). Following DeiT, data augmentation includes standard random resized crop, horizontal flip, RandAugment \cite{randaugment}, Mixup \cite{mixup}, CutMix \cite{cutmix}, Random Erasing \cite{random_erasing} and color jitter. For regularization, label smoothing \cite{inception_v3}, stochastic depth \cite{stochastic_depth}, and weight decay are adopted. Like ConvNeXt, we also use LayerScale \cite{layerscale}, a technique to help train deep models. Our code is based on PyTorch \cite{pytorch} and timm \cite{rw2019timm}.

\begin{table}[t]
\begin{center}
\footnotesize
\setlength{\tabcolsep}{3pt}
\begin{tabular}{l | c c c c c }
\whline
\multirow{2}{*}{\makecell[c]{Model}} &  \multirow{2}{*}{\makecell[c]{Params \\ (M)}} & \multirow{2}{*}{\makecell[c]{MACs \\ (G)}}  & \multicolumn{2}{c}{Throughput (img/s)} & \multirow{2}{*}{\makecell[c]{Top-1 \\ (\%)}} \\
~ & ~ & ~ & Train & Infer  & ~  \\
\whline
DeiT-S \cite{deit} & 22 & 4.6 & 276 & 784 & 79.8 \\ 
MetaNeXt-Attn & 22 & 4.6 & 288 & 816 & 3.9  \\ 
ConvNeXt-S (\textit{iso.}) \cite{convnext} & 22 & 4.3 & 270 & 879 &  79.7 \\
\modelname{}-S (\textit{iso.}) & 22 & 4.2 & 310 & 998 & 79.7 \\
\whline
\end{tabular}
\end{center}
\vspace{-3mm}
\caption{\textbf{Comparison among ViT, isotropic ConvNeXt and \modelname{}.}  MetaNeXt-Attn is instantiated from MetaNeXt with token mixer of self-attention \cite{transformer}. The throughputs are measured on 2080Ti (PyTorch 1.8.1 and CUDA 10.2) with FP32. The batch size for throughput benchmarking is initially set as 128 and is reduced until the GPU can host. The better results of ``Channel First" and ``Channel Last"  memory layouts are reported.}
\label{tab:iso}
\vspace{-9mm}
\end{table}

\myPara{Results} 
We compare \modelname{} with various state-of-the-art models, including attention-based and convolution-based models. As can be seen in Table \ref{tab:imagenet}, \modelname{} achieves highly competitive performance as well as enjoys higher speed. 
\modelname{} consistently enjoys better accuracy-speed trade-off than ConvNeXt \cite{convnext}.
For example, \modelname{}-T not only surpasses ConvNeXt-T by 0.2\%, but also enjoys $1.6 \times$/$1.2 \times $ training/inference throughputs on A100 than ConvNeXts, similar to those of ResNet-50. That is to say, \modelname{}-T enjoys both ResNet-50's speed and ConvNeXt-T's accuracy. 
Moreover,  following Swin and ConvNeXt, we also finetuned the \modelname{}-B trained at the resolution of $224 \times 224$ to $384 \times 384$ for 30 epochs.  We can see that \modelname{}-B obtains higher train and inference throughputs than ConvNeXt-B while keeping competitive accuracy.

It is observed that the speed improvement is much more significant for the lightweight model size, and the improvement gradually becomes smaller when the model size scales up.  The reason is that  computation complexity of depthwise and Inception depthwise convolutions are linear to channel number, \ie, $\mathcal{O}(C)$ where $C$ is channel number. For MLPs, their computation complexity is
$\mathcal{O}(C^2)$. For larger models (larger $C$), its computation is further dominated by MLPs. By only improving depthwise convolution, the speed improvement becomes smaller when the model is larger.

Besides the 4-stage framework \cite{vgg, resnet, swin}, another notable one is ViT-style \cite{vit} isotropic architecture which has only one stage. To match the parameters and MACs of DeiT-S, we construct \modelname{}-S (\textit{iso.}) following ConvNeXt-S (\textit{iso.}) \cite{convnext}. Specifically, we set the embedding dimension as 384 and the block number as 18. Besides, we build a model called MetaNeXt-Attn which is instantiated from MetaNeXt block by specifying self-attention as token mixer. The aim of this model is to investigate whether it is possible to merge two residual sub-blocks of the Transformer block into a single one. The experiment results are shown in Table \ref{tab:iso}. It can be seen that \modelname{} can also perform well with the isotropic architecture, demonstrating \modelname{} exhibits good generalization across different frameworks. It is worth noting that MetaNeXt-Attn could not be trained to converge and only achieved an accuracy of 3.9\%. This result suggests that, unlike the token mixer in MetaFormer, the token mixer in MetaNeXt cannot be too complex. If it is, the model may not be trainable.

\begin{table}[t]
\centering
\footnotesize
\setlength{\tabcolsep}{6pt}
\begin{tabular}{l|cccc}
\whline
\multirow{2}{*}{Backbone} & \multicolumn{3}{c}{UperNet}\\
\cline{2-5}
& Params (M) & MACs (G) & FPS & mIoU (\%) \\
    \whline
    Swin-T~\cite{swin}                & 60 & 945 & 20.6 & 45.8 \\
    ConvNeXt-T ~\cite{convnext}          & 60 & 939 & 20.6 & 46.7 \\
    \modelname{}-T  & 56 & 933 & 22.7 & \textbf{47.9} \\
	\hline
    Swin-S~\cite{swin}                & 81 & 1038 & 16.2 & 49.5 \\
    ConvNeXt-S ~\cite{convnext}          & 82 & 1027 & 16.8 & 49.6 \\
    \modelname{}-S & 78 & 1020 & 17.6 & \textbf{50.0} \\
	\hline
    Swin-B~\cite{swin}                & 121 & 1188 & 16.2 & 49.7 \\
     ConvNeXt-B ~\cite{convnext}          & 122 & 1170 & 15.7 & 49.9 \\
    \modelname{}-B  & 115 & 1159 & 17.5 & \textbf{50.6} \\
\whline
\end{tabular}

\caption{\textbf{Performance of semantic segmentation with UperNet \cite{upernet} on ADE20K~\cite{ade20k} validation set.} Images are cropped to $512 \times 512$ for training. The MACs are measured with input size of $512 \times 2048$. The FPS are benchamrked on 2080Ti.}
\label{tab:upernet}
\vspace{-6mm}
\end{table}

\begin{table}[t]
\centering
\footnotesize
\setlength{\tabcolsep}{4pt}
\begin{tabular}{l|cccc}
\whline
\multirow{2}{*}{Backbone} & \multicolumn{3}{c}{Semantic FPN} \\
\cline{2-5}
& Params (M) & MACs (G) & FPS & mIoU (\%) \\
    \whline
    ResNet-50~\cite{resnet}                & 29 & 46 & 30.2 & 36.7 \\
    PVT-Small~\cite{pvt}          & 28 & 45 & 27.2 & 39.8 \\
	PoolFormer-S24  \cite{metaformer}                  & 23 & 39 & 28.8 & 40.3 \\
    \modelname{}-T & 28 & 44 & 31.4 & \textbf{43.1} \\
    \hline
    ResNet-101~\cite{resnet}      & 48 & 65 &  22.2 & 38.8\\
    ResNeXt-101-32x4d~\cite{resnext} & 47 & 65 & -- & 39.7 \\
    PVT-Medium~\cite{pvt}         & 48 & 61 & 20.0 & 41.6 \\
     PoolFormer-S36 \cite{metaformer}                   & 35 & 48 & 21.6 & 42.0 \\
    PoolFormer-M36 \cite{metaformer}                   & 60 & 68 & 15.4 & 42.4 \\
     \modelname{}-S & 50 & 65 & 20.7 & \textbf{45.6} \\
    \hline
    PVT-Large~\cite{pvt}          & 65 & 80 & 16.0 & 42.1 \\
    ResNeXt-101-64x4d~\cite{resnext} & 86 & 104 & -- & 40.2 \\
     PoolFormer-M48 \cite{metaformer}                  & 77 & 82 & 12.1 & 42.7 \\
     \modelname{}-B & 85 & 100 & 20.2 & \textbf{46.4} \\
\whline
\end{tabular}

\caption{\textbf{Performance of semantic segmentation with Semantic FPN \cite{fpn} on ADE20K~\cite{ade20k} validation set.} Images are cropped to $512 \times 512$ for training. The MACs are measured with input size of $512 \times 512$. The FPS are benchamrked on 2080Ti.}
\label{tab:fpn}
\vspace{-9mm}
\end{table}

\begin{table*}[t]
\footnotesize
\centering
\setlength{\tabcolsep}{9pt}
\scalebox{1.0}{\begin{tabular}{l|l|c c c c c }
\whline
	\multirow{2}{*}{\makecell[c]{Ablation}} & \multirow{2}{*}{\makecell[c]{Variant}} & \multirow{2}{*}{\makecell[c]{Params \\ (M)}} & \multirow{2}{*}{\makecell[c]{MACs \\ (G)}} & \multicolumn{2}{c}{Throughput} & \multirow{2}{*}{\makecell[c]{Top-1 \\ (\%)}} \\
 ~ & ~ & ~ & ~ & Train & Inference & ~ \\
\whline
Baseline & None (\modelname{}-T) & 28.1 & 4.2 & 901 & 2900 & 82.3 \\
\hline
\multirow{4}{*}{\makecell[l]{Branch}} &  Remove horizontal band kernel & 28.0 & 4.2 & 947 & 3093 & 81.9 \\
~ & Remove vertical band kernel & 28.0 & 4.2 & 954 & 3173 & 81.9 \\
~ & Remove small band kernel & 28.0 & 4.2 & 940 & 3004 & 82.0 \\
~ & horizontal and vertical band kernel in parallel $\rightarrow$ in sequence & 28.1 & 4.2 & 903 & 2971 & 82.1 \\
\hline
\multirow{3}{*}{\makecell[l]{Band \\ kernel size}}
~ & Band kernel size 11 $\rightarrow$ 7 & 28.0 & 4.2 & 905 & 2946 & 82.1 \\ 
~ & Band kernel size 11 $\rightarrow$ 9 & 28.1 & 4.2 & 904 & 2916 & 82.1 \\
~ & Band kernel size 11 $\rightarrow$ 13 & 28.1 & 4.2 &  896 & 2895 & 82.0  \\
\hline
\multirow{2}{*}{\makecell[l]{Convolution \\
branch ratio}}
~ & Conv. branch ratio $1/8 \rightarrow 1/4$ & 28.1 & 4.2 & 834 & 2499 & 82.2 \\
~ & Conv. branch ratio $1/8 \rightarrow 1/16$ & 28.0 & 4.2 &  936 & 3097 & 81.8 \\
\hline
Normalization & Batch Norm \cite{batch_norm} $\rightarrow$ Layer Norm \cite{layer_norm} & 28.1 & 4.2 & 721 & 2646 & 82.4 \\
\whline
\end{tabular}
}
\caption{\textbf{Ablation for \modelname{} on ImageNet-1K classification benchmark.} \modelname{}-T is utilized as the baseline for the ablation study. Top-1 accuracy on the validation set is reported. The throughputs are measured on an A100 GPU (PyTorch 1.13.0 and CUDA 11.7.1) with TF32 and batch size of 128.
}
\label{tab:ablation}
\vspace{-9mm}
\end{table*}

\subsection{Semantic segmentation}
\myPara{Setup} 
ADE20K~\cite{ade20k}, a commonly used scene parsing benchmark, is used to evaluate our models on semantic segmentation task. ADE20K includes 150 fine-grained semantic categories, containing twenty thousand and two thousand images in the training set and validation set, respectively.
The checkpoints trained on ImageNet-1K \cite{imagenet_cvpr} at the resolution of $224^2$ are utilized to initialize the backbones. Following Swin \cite{swin} and ConvNeXt \cite{convnext}, we firstly evaluate \modelname{} with UperNet \cite{upernet}. The models are trained with AdamW \cite{adamw} optimizer with learning rate of 6e-5 and batch size of 16 for 160K iterations. Following PVT \cite{pvt} and PoolFormer \cite{metaformer}, \modelname{} is also evaluated with Semantic FPN \cite{fpn}. 
In common practices~\cite{fpn,chen2017deeplab}, the batch size is 16 for the setting of 80K iterations. Following PoolFormer \cite{metaformer}, we increase the batch size to 32 and decrease the iterations to 40K to speed up training.  AdamW \cite{adam, adamw}  is adopted with a learning rate of 2e-4 and a polynomial decay schedule of 0.9 power. Our code is based on PyTorch \cite{pytorch} and mmsegmentation \cite{mmseg2020}.

\myPara{Results}
For segmentation with UpNet \cite{upernet}, the results are shown in Table \ref{tab:upernet}. As can be seen, 
\modelname{} consistently outperforms Swin \cite{swin} and ConvNeXt \cite{convnext} for different model sizes. In the method of Semantic FPN \cite{fpn} as shown in Table \ref{tab:fpn}, \modelname{} significantly surpasses other backbones, like PVT \cite{pvt} and PoolFormer \cite{metaformer}. These results show that \modelname{} also has a high potential for dense prediction tasks.

\subsection{Ablation studies}
We conduct ablation studies on ImageNet-1K \cite{imagenet_cvpr, imagenet_ijcv} using \modelname{}-T as baseline from the following aspects.

\myPara{Branch} Inception depthwise convolution includes four branches, three convolutional ones, and identity mapping. When removing any branch of horizontal or vertical band kernel, performance significantly drops from 82.3\% to 81.9\%, demonstrating the importance of these two branches. This is because these two branches with band kernels can enlarge the receptive field of the model. For the branch of small square kernel size of $3\times 3$, removing it can still achieve up to 82.0\% top-1 accuracy and bring higher throughput. This inspires us that if we attach more importance to the model speed, the simple version of \modelname{} without the square kernel of $3\times 3$ can be adopted.  For the band kernel, Inception v3 mostly equips them in a sequential way. We find that this assembling method can also obtain similar performance and even a little speed up the model. A possible reason is that PyTorch/CUDA may have optimized sequential convolutions well, and we only implement the parallel branches at a high level (see Algorithm \ref{alg:code}). We believe the parallel method will be faster when it is optimized better. Thus, parallel method for the band kernels is adopted by default.

\myPara{Band kernel size} It is found the performance can be improved from kernel size 7 to 11, but it drops when the band kernel size increases to 13. This phenomenon may result from the optimization  and can be solved by methods like structural re-parameterization \cite{repvgg, replknet}. For simplicity, we  set the  kernel size as 11 by default except for atto size.

\myPara{Convolution branch ratio} When the ratio increases from $1/8$ to $1/4$, performance improvement can not be observed. Ma \etal \cite{shufflenet_v2} also point out that it is not necessary for all channels to conduct convolution. But when the ratio decreases to $1/16$, it brings a serious performance drop. This is because a smaller ratio would limit the degree of token mixing, resulting in performance drop. We thus set the convolution branch ratio as $1/8$ by default except for atto size.

\myPara{Normalization} When replacing the Batch Normalization \cite{batch_norm} with Layer Normalization \cite{layer_norm}, the performance improvement improve by 0.1\% but suffer throughput drop in both training and inference. Since this paper focuses on efficiency, we adopt Batch Normalization for \modelname{}.

\vspace{-1mm}
\section{Conclusion}
\vspace{-1mm}
In this work, we propose an effective and efficient CNN architecture, InceptionNeXt, which enjoys a better trade-off between the practical speed and the performance than previous network architectures. 
InceptionNeXt decomposes large-kernel depthwise convolution along channel dimension into four parallel branches, including identity mapping, a small square kernel, and two orthogonal band kernels. All these four branches are much more computationally efficient than a large-kernel depthwise convolution in practice, and can also work together to have a large spatial receptive field for good performance. Extensive experimental results demonstrate the superior performance and the high practical efficiency of InceptionNeXt.

{\small
\myPara{Acknowledgement}
This project is supported by the
National Research Foundation Singapore under its Medium Sized Center for Advanced Robotics Technology Innovation,
and the Advanced Research and Technology Innovation Centre (ARTIC), the National University of Singapore under Grant (project number: A0005947-21-00, project reference: ECT-RP2).
Weihao Yu was partly supported by Snap Research Fellowship, Google's TPU Research Cloud (TRC), and Google Cloud Research Credits program.
Pan Zhou was supported by the Singapore Ministry of Education (MOE) Academic Research Fund (AcRF) Tier 1 grant. We would like to thank Ross Wightman for integrating the model and code into Hugging Face's pytorch-image-models repository.
}

\clearpage
\setcounter{page}{1}
\maketitlesupplementary
\appendix

\section{Hyper-parameters}
\subsection{ImageNet-1K image classification}
On ImageNet-1K \cite{imagenet_cvpr, imagenet_ijcv} classification benchmark, following ConvNeXt \cite{convnext} and ConvNeXt-A trained by timm \cite{rw2019timm}, we adopt the hyper-parameters shown in Table \ref{tab:hyperparameter} to train \modelname{} at the input resolution of $224^2$ and fine-tune it at $384^2$. Our code is implemented by PyTorch \cite{pytorch} based on timm library \cite{rw2019timm}.

\subsection{Semantic segmentation}
For ADE20K \cite{ade20k} semantic segmentation, we utilize ConvNeXt as the backbone with UpNet \cite{upernet} following the configs of Swin \cite{swin}, and FPN \cite{fpn} following the configs of PVT \cite{pvt} and PoolFormer \cite{metaformer}. The backbone is initialized by checkpoints pre-trained on ImageNet-1K at the resolution of $224^2$. The peak stochastic depth rates of the \modelname{} backbone are shown in Table \ref{tab:hyperparameter_ade}. Our implementation is based on PyTorch \cite{pytorch} and mmsegmentation library \cite{mmseg2020}.

\section{Qualitative results}
\begin{figure*}[h]
    \centering
    \begin{subfigure}[b]{0.19\textwidth}
        \centering
        \includegraphics[width=1\textwidth]{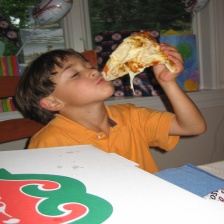}
        \includegraphics[width=1\textwidth]{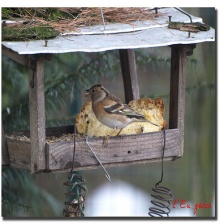}
        \includegraphics[width=1\textwidth]{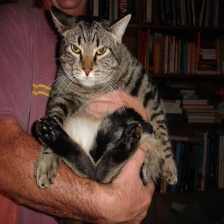}
        \includegraphics[width=1\textwidth]{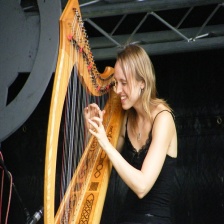}
    \end{subfigure}    
    \begin{subfigure}[b]{0.19\textwidth}
        \centering
        \includegraphics[width=1\textwidth]{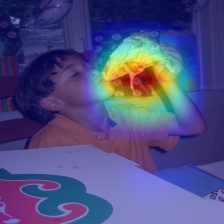}
        \includegraphics[width=1\textwidth]{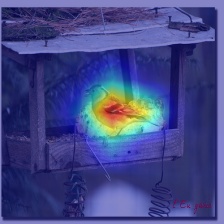}
        \includegraphics[width=1\textwidth]{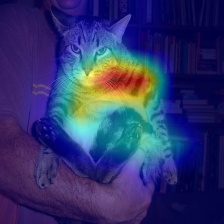}
        \includegraphics[width=1\textwidth]{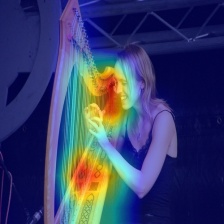}
    \end{subfigure}  
    \begin{subfigure}[b]{0.19\textwidth}
        \centering
        \includegraphics[width=1\textwidth]{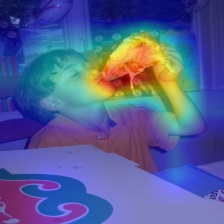}
        \includegraphics[width=1\textwidth]{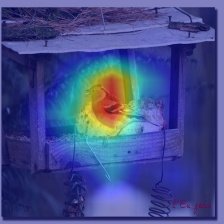}
        \includegraphics[width=1\textwidth]{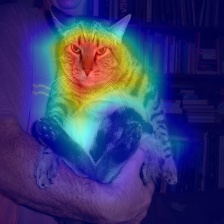}
        \includegraphics[width=1\textwidth]{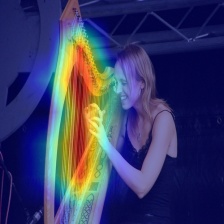}
    \end{subfigure}  
    \begin{subfigure}[b]{0.19\textwidth}
        \centering
        \includegraphics[width=1\textwidth]{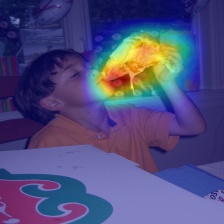}
        \includegraphics[width=1\textwidth]{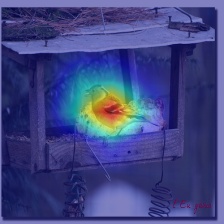}
        \includegraphics[width=1\textwidth]{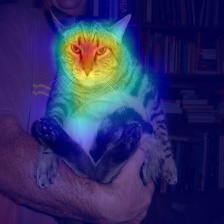}
        \includegraphics[width=1\textwidth]{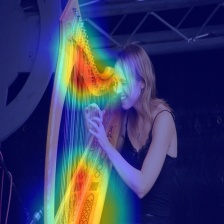}
    \end{subfigure}  
    \begin{subfigure}[b]{0.19\textwidth}
        \centering
        \includegraphics[width=1\textwidth]{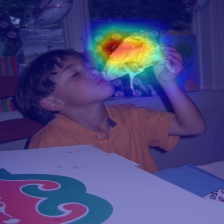}
        \includegraphics[width=1\textwidth]{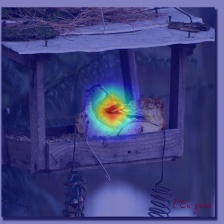}
        \includegraphics[width=1\textwidth]{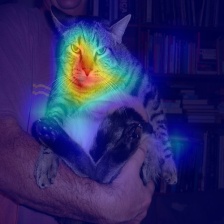}
        \includegraphics[width=1\textwidth]{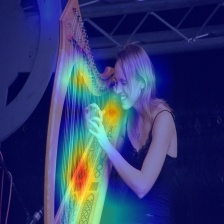}
    \end{subfigure}  
    \begin{center}
    	 ~~~~~~~Input\qquad \qquad \quad RSB-ResNet-50 \cite{resnet, resnetsb} \qquad  Swin-T \cite{deit} \qquad ~~~~ ConvNeXt-T \cite{convnext} \qquad ~~ \modelname{}-T
    \end{center}  
    \caption{
        \label{fig:grad_cam} Grad-CAM \cite{gradcam} activation maps of different models trained on ImageNet-1K. The visualized images are from the validation set of ImageNet-1K. 
    }
\end{figure*}

Grad-CAM \cite{gradcam} is employed to visualize the activation maps of different models trained on ImageNet-1K, including RSB-ResNet-50 \cite{resnet, resnetsb}, Swin-T \cite{swin}, ConvNeXt-T \cite{convnext} and our \modelname{}-T. The results are shown in Figure \ref{fig:grad_cam}. Compared with other models, \modelname{}-T locates key parts more accurately with smaller activation areas.

\begin{table}[h]
\setlength{\tabcolsep}{4pt}
\footnotesize
\centering
\begin{tabular}{@{}l|c|ccc|c@{}}
\whline
 & \multicolumn{5}{c}{\modelname{}} \\ 
\cline{2-6}
 & \multicolumn{4}{c|}{Train} & Finetune \\
 & \multicolumn{1}{c}{A} & T & S & B & B \\
\whline
Input resolution & $224^2$ & \multicolumn{3}{c|}{$224^2$} & $384^2$ \\
Epochs & 450 & \multicolumn{3}{c|}{300} & 30 \\
Batch size & 1280 & \multicolumn{3}{c|}{4096} & 1024 \\
Optimizer & AdamW & \multicolumn{3}{c|}{AdamW} & AdamW \\
Adam $\epsilon$ & 1e-8 & \multicolumn{3}{c|}{1e-8} & 1e-8 \\
Adam $(\beta_1, \beta_2)$ & (0.9, 0.999) & \multicolumn{3}{c|}{(0.9, 0.999)} & (0.9, 0.999) \\
Learning rate & 1e-3 & \multicolumn{3}{c|}{4e-3} & 5e-5 \\
Learning rate decay & Cosine &  \multicolumn{3}{c|}{Cosine} & Cosine \\
Gradient clipping & None & \multicolumn{3}{c|}{None} & None \\
Warmup epochs & 5 & \multicolumn{3}{c|}{20} & None \\
Weight decay & 0.05 & \multicolumn{3}{c|}{0.05} & 0.05 \\
Rand Augment & 5/uniform & \multicolumn{3}{c|}{9/0.5} & 9/0.5 \\
Repeated Augmentation & off & \multicolumn{3}{c|}{off} & off \\
Cutmix & 1.0 & \multicolumn{3}{c|}{1.0} & 1.0 \\
Mixup & 0.2 & \multicolumn{3}{c|}{0.8} & 0.8 \\
Cutmix-Mixup switch prob & 0.5 & \multicolumn{3}{c|}{0.5} & 0.5 \\
Random erasing prob & 0.1 & \multicolumn{3}{c|}{0.25} & 0.25 \\
Label smoothing & 0.1 & \multicolumn{3}{c|}{0.1} & 0.1 \\
Peak stochastic depth rate & 0.1 & 0.1 & 0.3 & 0.4 & 0.7 \\
Dropout in classifier & 0.0 & \multicolumn{3}{c|}{0.0} & 0.5 \\
LayerScale initialization & 1e-6 & \multicolumn{3}{c|}{1e-6}  & Pre-trained \\
Random erasing prob & 0.1 & \multicolumn{3}{c|}{0.25} & 0.25 \\
EMA decay rate & None & \multicolumn{3}{c|}{None} & 0.9999 \\
\whline
\end{tabular}

\caption{\textbf{Hyper-parameters of \modelname{} on ImageNet-1K image classification.}
\label{tab:hyperparameter}
}
\end{table}

\begin{table}[h]
\centering
\begin{tabularx}{0.5\textwidth}{@{}l|YYY}
\whline
\multirow{2}{*}{\makecell[c]{Method}} & \multicolumn{3}{c}{\modelname{} stochastic depth rate} \\ 
\cline{2-4}
 &  T  &  S  &  B  \\
\whline
UperNet \cite{upernet} & 0.2 & 0.3 & 0.4 \\
FPN \cite{fpn} & 0.1 & 0.2 & 0.2 \\
\whline
\end{tabularx}

\caption{\textbf{Stochasic depth rate of \modelname{} backbone with UperNet and FPN for ADE20K semantic segmentation.}
\label{tab:hyperparameter_ade}
}
\end{table}

{
    \small
    	 
  \bibliographystyle{ieeenat_fullname}
    \bibliography{main}
}

\end{document}